\documentclass[a4paper,UKenglish]{article}

\usepackage{amsmath}
\usepackage{amsthm}
\usepackage{amssymb}
\usepackage{mathtools}
\usepackage{hyperref}
\usepackage{color}
\usepackage{algorithm}
\usepackage[noend]{algpseudocode}
\usepackage{microtype}

\makeatletter
\@ifclassloaded{lipics-v2016}{}{
	\usepackage{authblk}
	\usepackage[capitalize,noabbrev]{cleveref}

	\theoremstyle{plain}
	\newtheorem{theorem}{Theorem}
	\newtheorem{lemma}[theorem]{Lemma}

	\theoremstyle{definition}
	\newtheorem{definition}[theorem]{Definition}
	\newtheorem{example}[theorem]{Example}
}
\makeatother

\theoremstyle{plain}
\newtheorem{claim}[theorem]{Claim}

\theoremstyle{definition}
\newtheorem{observation}[theorem]{Observation}

\theoremstyle{plain}

\newcommand{\BE}{\begin{enumerate}}
\newcommand{\EE}{\end{enumerate}}
\newcommand{\BI}{\begin{itemize}}
\newcommand{\EI}{\end{itemize}}
\newcommand{\I}{\item}
\newcommand{\BA}{\begin{aligned}}
\newcommand{\EA}{\end{aligned}}
\newcommand{\BPF}{\begin{proof}}
\newcommand{\EPF}{\end{proof}}

\newcommand{\B}[1]{\left( {#1} \right)}
\newcommand{\F}[2]{{\frac{#1}{#2}}}
\newcommand{\R}[1]{{\frac{1}{#1}}}
\newcommand{\BF}[2]{\B{\F{#1}{#2}}}

\newcommand{\set}[1]{\left\{ #1 \right\}}
\newcommand{\midline}[2]{{#1} \, \middle| \, {#2}}
\newcommand{\prob}[1]{\text{\tt Pr}\left[ #1 \right]}
\newcommand{\sexpct}[2]{\text{\tt E}_{{#1}}\left[#2\right]}

\newcommand{\expct}[1]{\sexpct{}{#1}}
\newcommand{\stset}[2]{\set{\midline{#1}{#2}}}
\newcommand{\cprob}[2]{\prob{\midline{#1}{#2}}}
\newcommand{\cexpct}[2]{\expct{\midline{#1}{#2}}}

\newcommand{\Floor}[1]{{\left\lfloor {#1} \right\rfloor}}
\newcommand{\Ceil}[1]{{\left\lceil {#1} \right\rceil}}

\newcommand{\Natural}{\mathbb{N}}

\newcommand*\diff{\mathop{}\!\mathrm{d}}

\definecolor{antiquefuchsia}{rgb}{0.57, 0.36, 0.51}

\newcommand{\RIGHT}{\quad \Longrightarrow \quad}

\def\YesInside{Y}

\def\Inside{N}

\renewcommand{\[}[1][]{
	\def\temp{#1} 
	\ifx\temp\empty
		\begin{equation*} 
		\def\Inside{Y}
	\else 
		\begin{equation}\label{#1}
	\fi
}
\renewcommand{\]}{
\ifx\Inside\YesInside
	\def\Inside{N}
	\end{equation*}
\else
	\end{equation}
\fi
}

\newcommand{\repeatable}[5]{
\begin{#1}{\label{#4}}
{#5}
\end{#1}
\theoremstyle{#3}
\newtheorem*{temp#4}{#2 \ref{#4}}
\expandafter\newcommand\csname #4\endcsname{
\begin{temp#4}
{#5}
\end{temp#4}}
}

\newcommand{\rLemma}[2]{\repeatable{lemma}{Lemma}{plain}{#1}{#2}}
\newcommand{\rClaim}[2]{\repeatable{claim}{Claim}{plain}{#1}{#2}}
\newcommand{\rExample}[2]{\repeatable{example}{Example}{definition}{#1}{#2}}

\usepackage[margin=1in]{geometry}

\newcommand{\cC}{{\cal C}}
\newcommand{\cI}{{\cal I}}
\newcommand{\cX}{{\cal X}}
\newcommand{\cA}{{\cal A}}
\newcommand{\cB}{{\cal B}}

\newcommand{\Ri}{\set{1,\ldots,i-1}}
\newcommand{\Time}{{\mathbb T}}
\newcommand{\Price}{\texttt{Price}}

\newcommand{\state}{\delta}
\newcommand{\simple}{A_{\text{\tt simp}}}
\newcommand{\nfin}{\texttt{SC}}

\title{The Dependent Doors Problem: An Investigation into Sequential Decisions without Feedback\footnote{This work has received funding from the European Research Council (ERC) under the European Union's Horizon 2020 research and innovation programme (grant agreement No 648032).}}

\author[1]{Amos Korman}
\author[2]{Yoav Rodeh}
\affil[1]{CNRS and University Paris Diderot\\
  Paris, France\\
  \texttt{amos.korman@irif.fr}}
\affil[2]{Weizmann Institute of Science\\
  Rehovot, Israel\\
  \texttt{yoav.rodeh@gmail.com}}

\begin{document}
\maketitle

\begin{abstract}
We introduce the {\em dependent doors problem} as an abstraction for situations in which one must perform a sequence of possibly dependent decisions, without receiving  feedback information on the effectiveness of previously made actions. 
Informally, the problem considers a set of $d$ doors that are initially closed, and the aim is to open all of them as fast as possible.
To open a door, the algorithm knocks on it and it might open or not according to some probability distribution. 
This distribution may depend on which other doors are currently open, as well as on which other doors were open during each of the previous knocks on that door. 
The algorithm aims to minimize the expected time until all doors open. Crucially, it must act at any time without knowing whether or which other doors have already opened. 
In this work, we focus on scenarios where  dependencies between doors are both positively correlated and acyclic.

The fundamental distribution of a door describes the probability it opens in the best of conditions (with respect to other doors being open or closed). We show that if in two configurations of $d$ doors corresponding doors share the same fundamental distribution, then these configurations have the same optimal running time up to a universal constant, no matter what are the dependencies between doors and what are the distributions. 
We also identify algorithms that are optimal up to a universal constant factor. For the case in which all doors share the same fundamental distribution we additionally provide a simpler algorithm, and a formula to calculate its running time.
We furthermore analyse the price of lacking feedback for several configurations governed by standard fundamental distributions. In particular, we show that the price is logarithmic in $d$ for memoryless doors, but can potentially grow to be linear in $d$ for other distributions.

We then turn our attention to investigate precise bounds. Even for the case of two doors, identifying the optimal sequence is an intriguing combinatorial question. 
Here, we study the case of two cascading  memoryless  doors. That is, the first door opens on each knock  independently with probability $p_1$. The second door can only open if the first door is open, in which case it will open on each knock independently with probability $p_2$. We solve this problem almost completely by identifying algorithms that are optimal up to an additive term of 1. 
 
\end{abstract} 

\section{Introduction} \label{sec:introduction}


%
%
%
Often it is the case that one must accomplish multiple tasks whose success probabilities are dependent on each other. In many cases, failure to achieve one task will tend to have a more negative affect on the success probabilities of other tasks. In general, such dependencies may be quite complex, and balancing the work load  between different tasks becomes a computational challenge. The situation is further complicated if the ability to detect whether a task has been accomplished is limited. 
For example, if task $B$ highly depends on task $A$ then until $A$ is accomplished, all efforts invested in $B$ may be completely wasted. How should one divide the effort between these tasks if feedback on the success of $A$ is not available? 

In this preliminary work we propose a setting that captures some of the  fundamental  challenges that are inherent to the process of decision making without feedback. 
We introduce the {\em dependent doors} problem, informally described as follows. There are $d \geq 2$ doors (representing tasks) which are initially closed, and the aim is to open all of them as fast as possible. 
To open a door, the algorithm can ``knock'' on it and it might open or not according to some governing probability distribution, that may depend on  other doors being open or closed\footnote{Actually, the distribution associated with some door $i$ may depend on the state of other doors (being open or closed) not only at the current knock, but also at the time of each of the previous knocks on door $i$.}. We focus on settings in which doors are positively correlated, which informally means that the probability of opening a door is never decreased if another door is open.
The governing distributions and their dependencies are known to the algorithm in advance. Crucially, however, during the execution, it gets no direct feedback on whether or not a door has opened unless all $d$ doors have opened, in which case the task is completed. 

This research has actually originated from our research on heuristic search on trees \cite{FOCS}. Consider a tree of depth $d$ with a treasure placed at one of its leaves. At each step the algorithm can ``check'' a vertex, which is child of an already checked vertex. Moreover, for each level of the tree, the algorithm has a way to compare the previously checked vertices on that level. This comparison has the property that if the ancestor of the treasure on that level was already checked, then it will necessarily be considered as the ``best'' on that level. Note, however, that unless we checked all the vertices on a given level, we can never be sure that the vertex considered as the best among checked vertices in the level 
is indeed the correct one. With such a guarantee, and assuming that the algorithm gets no other feedback from checked vertices, any reasonable algorithm that is about to check a vertex on a given level, will always choose to check a child of the current best vertex on the level above it. Therefore, the algorithm can be described as a sequence of levels to inspect. Moreover, if we know the different distributions involved, then we are exactly at the situation of the dependent doors problem. See Appendix \ref{app:example} for more details on this example. 

Another manifestation of $d$ dependent doors can arise in the context of cryptography. Think about a sequence of $d$ cascading encryptions, and separate decryption 
protocols to attack each of the encryptions. Investing more efforts in decrypting the $i$'th encryption would increase the chances of breaking it, but only if previous encryptions where already broken. On the other hand, we get no feedback on an encryption being broken unless all of them are.

The case of two doors can serve as an abstraction for exploration vs.\ exploitation problems, where it is typically the case that deficient performances on the exploration part may result in much waste on the exploitation part \cite{Multi-Armed,Reinforced}. 
It can also be seen as the question of balance between searching and verifying in algorithms that can be partitioned thus \cite{STOC2013,Montgomery}. In both examples, there may be partial or even no feedback in the sense that we don't know that the first procedure succeeded unless the second one also succeeds. 

For simplicity, we concentrate on scenarios in which the dependencies  are {\em acyclic}. That is, if we draw the directed dependency graph between doors, then this graph does not contain any directed cycles. The examples of searching and verifying and the heuristic search on trees can both be viewed as acyclic. Moreover, despite the fact that many configurations are not purely acyclic, one can sometimes obtain a useful approximation that is.

To illustrate the problem, consider the following presumably simple case of two dependent memoryless doors. The first door opens on each knock  independently with probability $1/2$. The second door can only open if the first door is open, in which case it opens on each knock independently, with probability $1/2$. What is the sequence of knocks that minimizes the expected time to open both doors, remembering that we don't know when door 1 opens? It is easy to see that the alternating sequence $1,2,1,2,1,2,\ldots$ results in $6$ knocks in expectation. Computer simulations indicate that the best sequence gives a little more than $5.8$ and starts with $1, 1, 2, 2, 1, 2, 1, 2, 2, 1, 2, 2, 1, 2, 1, 2$. Applied to this particular scenario, our theoretical lower bound gives $5.747$, and our upper bound gives a sequence with expected time $5.832$.

\subsection{Context and Related Work}

This paper falls under the framework of decision making under uncertainty, a large research subject that has received significant amount of attention from researchers in various disciplines, including computer science, operational research, biology, sociology, economy, and even psychology and cognition, see, {\em e.g.,}  \cite{OperationsResearch,Benor,psycology,STOC2016,Feige94,giraldeau2000social,Karp07,Pelc02}. 
 
 Performing despite limited feedback would fit the framework of reinforced learning \cite{Reinforced} and is inherent to the study of exploration vs.~exploitation type of problems, including Multi-Armed Bandit problems~\cite{Multi-Armed}. In this paper we study the impact of having no feedback whatsoever. Understanding this extreme 
 scenario may serve as an approximation for cases where feedback is highly restricted, or limited in its impact. For example, if it turns out that the price of lacking feedback is small, then it may well be worth to avoid investing efforts in complex methods for utilizing the partial feedback.

%

Of particular interest is the case of two doors. As mentioned, difficulties resulting from the lack of feedback can arise when one aims to find  a solution by alternating between two subroutines: Producing promising candidate solutions  and verifying these candidates. Numerous  strategies are based on this interplay, including heuristics  based on brute force or trail and error approaches \cite{STOC2013,Montgomery}, sample and predict approaches \cite{Vazirani,Mitchell,Reinforced}, iterative local algorithms \cite{DC,Luby86}, and many others. Finding  strategies for efficiently balancing these two tasks can be therefore applicable.

\subsection{Setting}\label{sec:setting}

There are $d \geq 2$ doors and each door can be either open or closed.
Doors start closed, and once a door opens it never closes.
To open a door, an algorithm can knock on it and it might open or not according to some probability distribution. The goal is to minimize the expected number of knocks until all doors  open. Crucially, the algorithm has no feedback on whether or not a door has opened, unless all doors have opened, in which case the task is completed. 

The probability that a door opens may depend on the state of other doors (being open or closed) at the time of the current knock as well as on their state during each of the previous knocks on the door. For example, the probability that a certain knock at door $i$ succeeds may depend on the number of previous knocks on door $i$, but counting only those that were made while some other specific door $j$ was open. 
The idea behind this definition is that the more time we invest in opening a door the more likely it is to open, and the quality of each knock depends on what is the state of the doors it depends on at the time of the knock. 

Below we provide a semi-formal description of the setting. The level of detail is sufficient to understand the content of the main text, which is mainly concerned with independent and cascading configurations. The reader interested in a more formal description of the model is deferred to Appendix \ref{apx:general}.

A specific setting of doors is called a {\em configuration} (normally denoted $\cC$). This includes a description of all dependencies between doors and the resulting probability distributions. 
In this paper we assume that the dependency graph of the doors is acyclic, and so we may assume that a configuration describes an ordering of the doors, such that each door depends only on lower index doors.
Furthermore, we assume that the correlation between doors is positive, {\em i.e.}, a door being open can only improve the chances of other doors to open. 

Perhaps the simplest configuration is when all doors are {\em independent} of each other. In this case, door $i$ can be associated with a function $p_i : \Natural \rightarrow [0,1]$, where $p_i(n)$ is the probability that door $i$ is not open after knocking on it $n$ times. 
Another family of acyclic configurations are  {\em cascading} configurations. Here, door $i$ cannot open unless all doors of lower index are already open. 
In this case, the configuration can again be described by a set of functions $\{p_i\}_{i=1}^d$, where $p_i(n)$  describes the probability that door $i$ is not open after knocking on it $n$ times, where the count starts only after door $i-1$ is already open.

In general, given a configuration, each door $i$ defines a non-decreasing function $p_i : \Natural \rightarrow [0,1]$, called the
{\em fundamental distribution} of the door, where  $p_i(n)$ is the probability that the door is not open after knocking on it $n$ times in the best of conditions, {\em i.e.}, assuming all doors of lower index are open. In the case of independent and cascading configurations, the fundamental distribution $p_i$ coincides with the functions mentioned above.
Two doors are {\em similar} if they have the same fundamental distribution.
Two configurations are {\em similar} if for every $i$, door $i$ of the first configuration is similar to door $i$ of the second. 

When designing an algorithm, we will assume that the configuration it is going to run in is known.
As there is no feedback, a deterministic algorithm can be thought of as a possibly infinite sequence of door knocks.
A randomized algorithm is therefore a distribution over sequences, and as all of them will have expected running time at least as large as that of an optimal sequence (if one exists), the expected running time of a randomized algorithm cannot be any better. 
Denote by $\Time_\cC(\pi)$, the expected time until all doors open when running sequence $\pi$ in configuration $\cC$. 
We define $\Time_\cC = \min_\pi \Time_\cC(\pi)$. 
By Claim \ref{clm:optimal} in Appendix \ref{apx:existence}, there exists a sequence achieving this minimum. 
Therefore, by the aforementioned arguments, we can restrict our discussion to deterministic algorithms only.

If we had feedback we would knock on each door until it opens, and then continue to the next. 
Denoting by $E_i = \sum_{n=0}^\infty p_i(n)$ the expected time to open door $i$ on its own, 
the expected running time then does not depend on the specific dependencies between doors at all, and is $\sum_i E_i$. Also, this value is clearly optimal. To evaluate the impact of lacking feedback for a configuration $\cC$, we therefore define:
\[
\Price(\cC) = \F{\Time_\cC}{\sum_i E_i}
\]
Obviously  $\Price(\cC) \geq 1$, and for example, if all doors start closed and open after just $1$ knock, it is in fact equal to $1$. Claim \ref{clm:upperPrice} in Appendix \ref{apx:upperPrice} shows that $\Price(\cC) \leq d$.

\subsection{Our Results}
We have two main results. 
The first one, presented in Section \ref{sec:multiple}, states that any two similar configurations have the same optimal running time up to a constant factor. We stress that this constant factor is universal in the sense that it does not depend on the specific distributions or on the number of doors $d$. 
 
Furthermore, given a configuration, we identify an algorithm that is optimal for it up to a constant factor. We then show that for configurations where all doors are similar, there is a much simpler  algorithm which is optimal up to a constant factor, and describe a formula that computes its approximate running time.
We conclude Section \ref{sec:multiple} by analysing the price of lacking feedback for several configurations governed by standard fundamental distributions. In particular, we show that the price is logarithmic in $d$ for memoryless doors, but can potentially grow to be linear in $d$ for other distributions.

We then turn our attention to identify exact optimal sequences. Perhaps the simplest case is the case of two cascading memoryless doors. That is, the first door opens on each knock  independently with probability $p_1$. The second door can only open if the first door is open, in which case it opens on each knock independently, with probability $p_2$. 
In Section \ref{sec:geometric} we present our second main result: 
Algorithms for these  configurations that achieve the precise optimal running time up to an additive term of 1. 

On the technical side, to establish such an extremely competitive algorithm, we first consider a semi-fractional variant of the problem and find a sequence that achieves the precise optimal bound.
We then approximate this semi-fractional sequence to obtain an integer solution losing only an additive term of 1 in the running time. 
A nice anecdote is that in the case where $p_1 = p_2$ and are very small, the ratio of 2-knocks over 1-knocks in the sequence we get approaches the golden ratio.
Also, in this case, the optimal running time  approaches $3.58/p_1$ as $p_1$ goes to zero. It follows that in this case, the price of lacking feedback tends to $3.58/2$ and the price of dependencies, {\em i.e.}, the multiplicative gap between the cascading and independent settings, tends to $3.58/3$.

\section{Near Optimal Algorithms}\label{sec:multiple}

The following important lemma is proved in Appendix \ref{sec:general}  using a coupling argument:
\rLemma{extreme}{
Consider similar configurations $\cC, \cX$ and $\cI$, where $\cX$ is cascading and $\cI$ is independent. For every sequence $\pi$, 
$\Time_\cI(\pi) \leq \Time_\cC(\pi) \leq \Time_\cX(\pi)$.
This also implies that 
$\Time_\cI \leq \Time_\cC \leq \Time_\cX$.
}
The next theorem presents a near optimal sequence of knocks for a given configuration. 
In fact, by Lemma \ref{extreme}, this sequence is near optimal for any similar configuration, and so we get that the optimal running time for any two similar configurations is the same up to a universal multiplicative factor.
\begin{theorem} \label{thm:main1}
There is a polynomial algorithm\footnote{A polynomial algorithm in our setting
generates the next knock in the sequence in polynomial time in the index of the knock and in $d$, assuming that reading any specific value of any of the fundamental distributions of a door takes constant time.}, that given a configuration $\cC$ generates a sequence $\pi$ such that $\Time_\cC(\pi) = \Theta(\Time_\cI)$. In fact, $\Time_\cC(\pi) \leq 2+4\Time_\cI \leq 2+4\Time_\cC$. 
\end{theorem}
\BPF
Denote by $p_1, \ldots, p_d$ the fundamental distributions of the doors of $\cC$.
For a finite sequence of knocks $\alpha$, 
denote by $\nfin_\cC(\alpha)$ the probability that after running $\alpha$ in configuration $\cC$, some of the doors are still closed. 
Note that if $\alpha$ is {\em sorted}, that is, if all knocks on door 1 are done first, followed by the knocks on doors 2, etc., then $\nfin_\cX(\alpha) = \nfin_\cI(\alpha)$.

We start by showing that for any $T$, we can construct in polynomial time a finite  sequence $\alpha_T$ of length $T$ that maximizes the probability that all doors will open, 
{\em i.e.}, minimizes $\nfin_\cI(\alpha_T)$. As  noted above, if we sort the sequence, this is equal to $\nfin_\cX(\alpha_T)$.

The algorithm follows a dynamic programming approach, and calculates a matrix $A$, where $A[i, t]$ holds the maximal probability that a sequence of length $t$ has of opening all of the doors $1,2, \ldots, i$. 
All the entries $A[0,\cdot]$ are just $1$, and the key point is that for each $i$ and $t$, knowing all of the entries in $A[i,\cdot]$, it is easy to calculate $A[i+1,t]$:
\[
A[i+1,t] = \max_{k=0}^t A[i,t-k] \cdot \B{1 - p_{i+1}(k)}
\]
Calculating the whole table takes $O(dT^2)$ time, and $A[d,T]$ will give us the highest probability a sequence of length $T$ can have of opening all doors. Keeping tabs on the choices the $\max$ in the formula makes, we can get an optimal sequence $\alpha_T$, and can take it to be sorted.

Consider the sequence $\pi = \alpha_2 \cdot \alpha_4 \cdots \alpha_{2^n} \cdots$.
The complexity of generating this sequence up to place $T$ is $O(dT^2)$, and so this algorithm is polynomial.
Our goal will be to compare $\Time_\cX(\pi)$ with $\Time_\cI(\pi^\star)$, where $\pi^\star$ is the optimal sequence for $\cI$. 

The following observation stems from the fact that for any natural valued random variable $X$,  $\expct{X} = \sum_{n=0}^\infty \prob{X > n}$ and $\prob{X > n}$ is a non-increasing function of $n$.
\begin{observation} \label{obs:partial}
Let $\set{a_n}_{n=1}^\infty$ be a strictly increasing sequence of natural numbers, and $X$ be some natural valued random variable. Then:
\[
\sum_{n=1}^\infty (a_{n+1} - a_n) \prob{X > a_{n+1}}
\leq
\expct{X}
\leq
a_1 + \sum_{n=1}^\infty (a_{n+1} - a_n) \prob{X > a_n}
\]
\end{observation}
For a sequence $\pi$, denote by $\pi[n]$ the prefix of $\pi$ of length $n$. In this terminology, 
$\Time_\cC(\pi) = \sum_{n=0}^\infty \nfin_\cC(\pi[n])$.
Setting $a_n = 2 + 4 + \ldots + 2^n$ in the right side of Observation \ref{obs:partial}, and letting $X$ be the number of rounds until all doors open when using $\pi$, we get:
\[\BA
\Time_\cX(\pi) 
& \leq
2 + \sum_{n=1}^\infty 2^{n+1} \cdot \nfin_\cX(\pi[2 + \ldots + 2^n])
\leq
2 + \sum_{n=1}^\infty 2^{n+1} \cdot \nfin_\cX(\alpha_{2^n})
\\ & = 
2 + \sum_{n=1}^\infty 2^{n+1} \cdot \nfin_\cI(\alpha_{2^n})
\leq
2 + \sum_{n=1}^\infty 2^{n+1} \cdot \nfin_\cI(\pi^\star[2^n]) 
\leq 
2 + 4 \Time_\cI(\pi^\star)
\EA\]
The last step is using Observation \ref{obs:partial} with $a_n = 2^{n-1}$.
Theorem \ref{thm:main1} concludes.
\EPF

\subsection{Configurations where all Doors are Similar} \label{sec:equivalent}
In this section we focus on configurations where all doors have the same fundamental distribution $p(n)$. We provide simple algorithms that are optimal up to a universal constant, and establish the price of lacking feedback with respect to a few natural distributions.
Corresponding proofs appear in Appendix \ref{apx:price}.

\subsubsection{Simple Algorithms}
Let us consider the following  very simple algorithm $\simple$. It runs in phases, where in each phase it knocks on each door once, in order. As a sequence, we can write $\simple = (1, 2, \ldots, d)^\infty$.
Let $X_1, \ldots, X_d$ be i.i.d.\ random variables taking positive integer values, satisfying $\prob{X_i > n} = p(n)$. The following is straightforward:
\begin{claim} \label{clm:simpleMax}
$\Time_\cI(\simple) = \Theta\B{d \cdot \expct{ \max\set{X_1, \ldots, X_d}}}$
\end{claim} 
This one is less trivial:
\rClaim{clmAone}{
If all doors are similar  then $\Time_\cI(\simple) = \Theta(\Time_\cI)$
}
The claim above states that $\simple$ is optimal up to a multiplicative constant factor in the independent case, where all doors are similar. As a result, we can also show:
\rClaim{clmSimpleX}{
Denote by $\alpha_n$ the sequence $1^{2^n}, \ldots, d^{2^n}$. If all doors are similar then for any configuration $\cC$, 
$
\Time_\cC\B{\alpha_0 \cdot \alpha_1 \cdot \alpha_2 \cdots}
= \Theta(\Time_\cC)
$.
}
In plain words, the above claim states that the following algorithm is optimal up to a universal constant factor for any configuration where all doors are similar: Run in phases where phase $n$ consists of knocking $2^n$ consecutive times on each door, in order.

\subsubsection{On the Price of Lacking Feedback} \label{sub:price}

By Claims \ref{clm:simpleMax} and \ref{clmAone}, investigating the price of lacking feedback when all doors are similar boils down to understanding the expected maximum of i.i.d.\ random variables. 
\[[eq:priceByMax]
\Price = \Theta\BF{ \expct{\max\set{X_1, \ldots, X_d} }}{\expct{X_1}} 
\]
Note that we omitted dependency on the configuration, as by Theorem \ref{thm:main1}, up to constant factors, it is the same price as in the case where the doors are independent. 
Let us see a few examples of this value. First:
\rLemma{lmMax}{
If $X_1, \ldots, X_d$ are i.i.d.\ random variables taking natural number values, then:
\[
\expct{\max(X_1, \ldots, X_d)} = 
\Theta\B{\kappa + d\sum_{n=\kappa}^\infty \prob{X_i > n}}
\]
Where $\kappa = \min\stset{n \in \Natural}{\prob{X_1 > n} < 1/d}$
}

\begin{example}
After the first knock on it, each door opens with probability $1-1/d$ and if it doesn't, it will open at its $d+1$'st knock. The expected time to open each door on its own is $2$. By Lemma \ref{lmMax}, as $\kappa = d+1$, we get that $\Price = \Omega(\kappa) = \Omega(d)$. By Claim \ref{clm:upperPrice}, $\Price = \Theta(d)$.
\end{example}
\rExample{exGeometric}{
If $p(n) = q^n$ for some $1/2 < q < 1$, then 
$\Price = \Theta(\log(d))$.
}
\rExample{exPolynomial}{
If for some $c > 0$ and $a > 1$, $p(n) = \min(1, c/n^a)$, 
then $\Price = \Theta(d^\R{a})$.
}
Sometimes we know a bound on some moment of the distribution of opening a door.
If $\expct{X_1} < M$, then by Claim \ref{clm:upperPrice},
$\Time = O(d^2M)$. Also,
\rExample{exMoment}{
If $\expct{X_1^a} < M$ for some $a > 1$, then
$\Time = O\B{d^{1 + \R{a}} M^{1/a} (1 + \R{a-1})}$.
}
For example, if the second moment of the time to open a door on its own is bounded, we get an $O(d^{3/2})$ algorithm.

\section{Two Memoryless Cascading Doors}\label{sec:geometric}

One can say that by Theorem \ref{thm:main1} we solved much of the dependent doors problem. There is an equivalence of the independent and cascading models, and we give an up to constant factor optimal algorithm for any situation. 
However, we still find the question of finding the true optimal sequences for cascading doors to be an interesting one. What is the precise cost of having no feedback, in numbers? Even the simple case of two doors, each opening with probability $1/2$ on each knock, turns out to be quite challenging  and has a not so intuitive optimal sequence.

In this section, we focus on a very simple yet interesting case of the cascading door problem, and solve it almost exactly.
We have two doors. Door 1 opens with probability $p_1$ each time we knock on it, and  door 2 opens with probability $p_2$. We further extend the setting to consider different durations. Specifically, we assume that a knock on door 1 takes one time unit, and a knock on door 2 takes $c$ time units. Denote $q_1 = 1 - p_1$ and $q_2 = 1 - p_2$.
For brevity, we will call a knock on door 1 a {\em 1-knock}, and a knock on door 2 a {\em 2-knock}.

\subparagraph*{The Semi-Fractional Model.} As finding the optimal sequence directly proved to be difficult, we introduce a relaxation of our original model, termed the {\em semi-fractional model}.
In this model, we allow 1-knocks to be of any length. A knock of length $t$, where $t$ is a non-negative real number, will have probability of $1 - q_1^t$ of opening the door. 
In this case, a sequence consists of the alternating elements $1^t$ and $2$, where $1^t$ describes a knock of length $t$ on door 1.
We call sequences in the semi-fractional model {\em semi-fractional sequences}, and to differentiate, we call sequences in the original model {\em integer sequences}.

As our configuration $\cC$ will be clear from context, for a sequence $\pi$, we define $\expct{\pi} = \Time_\cC(\pi)$ to be the expected running time of the sequence. Clearly, every integer sequence has a similar semi-fractional sequence with the same expected running time.
As we will see, the reverse is not far from being true. 
That being so, finding the optimal semi-fractional sequence will give an almost optimal integer sequence.

\subsection{Equivalence of Models}

\begin{theorem} \label{thm:equiv2}
Every semi-fractional sequence $\pi$ has an integer sequence $\pi'$, s.t.,
$\expct{\pi'} \leq \expct{\pi} + 1$.
\end{theorem}
For this purpose, in this subsection only, we describe a semi-fractional sequence $\pi$ as a sequence of non-decreasing non-negative real numbers: $\pi_0, \pi_1, \pi_2, \ldots$, where $\pi_0 = 0$. This sequence describes the following semi-fractional sequence (in our original terms):
\[
1^{\pi_1 - \pi_0} \cdot 2 \cdot 1^{\pi_2 -\pi_1} \cdot 2 \cdots
\]
This representation simplifies our proofs considerably.
Here are some observations:
\BI
\I
1-knocks can be of length $0$, yet we still consider them in our indexing.
\I
The sequence is an integer sequence iff for all $i$, $\pi_i \in \Natural$.
\I
The $i$-th 2-knock starts at time $\pi_i + c(i-1)$ and ends at $\pi_i + ci$.
\I
The probability of door 1 being closed after the completion of the $i$-th 1-knock is $q_1^{\pi_i}$, and so the probability it opens at 1-knock $i$ is $q_1^{\pi_{i-1}} - q_1^{\pi_i}$
\EI
\begin{lemma} \label{lm:pushing}
For two sequences $\pi = (\pi_0, \pi_1, \ldots)$ and $\pi' = (\pi'_1, \pi'_2, \ldots)$,
if for all $i$, 
$\pi_i \leq \pi'_i \leq \pi_i + 1$
then $\expct{\pi'} \leq \expct{\pi} + 1$.
\end{lemma}

Lemma \ref{lm:pushing} is the heart of our theorem. Indeed, 
 once proven, Theorem \ref{thm:equiv2} follows in a straightforward manner. 
Given a semi-fractional sequence $\pi$, define $\pi'_i = \Ceil{\pi_i}$. Then, $\pi'$ is an integer sequence, and it satisfies the conditions of the lemma, so we are done. The lemma makes sense, as the sequence $\pi'$ in which for all $i>0$, $\pi_i' = \pi_i+1$, can be thought of as adding a 1-knock of length one in the beginning of the sequence. Even if this added 1-knock did nothing, the running time would increase by at most 1. However, the proof is more involved, since in the lemma, while some of the 2-knocks may have an increased chance of succeeding, some may actually have a lesser chance.
\BPF
Given  a sequence $\pi$ and an event $X$, we denote by $\cexpct{\pi}{X}$ the expected running time of $\pi$ given the event $X$.
Let $X_i$ denote the event that door 1 opens at its $i$-th 1-knock. As already said:
\[
\prob{X_i} 
= q_1^{\pi_{i-1}} - q_1^{\pi_i}
= \int_{\pi_{i-1}}^{\pi_i} q_1^x \ln(q_1) \diff x
\]
Where the last equality comes as no surprise, as it can be seen as modelling door 1 in a continuous fashion, having an exponential distribution fitting its geometrical one.
Now:
\[
\expct{\pi}
= 
\sum_{i=1}^\infty \prob{X_i} \cexpct{\pi}{X_i}
= 
\sum_{i=1}^\infty \int_{\pi_{i-1}}^{\pi_i} q_1^x \ln(q_1) \diff x \cdot \cexpct{\pi}{X_i} 
 = 
\int_0^\infty q_1^x \ln(q_1) \cdot \cexpct{\pi}{X_{i(x)}} \diff x
\]
Where $i(x) = \max_i\set{x \geq \pi_{i-1}}$, that is, the index of the 1-knock that $x$ belongs to when considering only time spent knocking on door 1. 
Defining $X'_i$ and $i'(x)$ in an analogous way for $\pi'$, we want to show that for all $x$,
\[
\cexpct{\pi'}{X'_{i'(x)}} \leq 1 + \cexpct{\pi}{X_{i(x)}}
\]
as using it with the last equality will prove the lemma. 
We need the following three claims:
\BE
\I
If $j \leq i$, then $\cexpct{\pi}{X_j} \leq \cexpct{\pi}{X_i}$
\I
For all $x$, $i'(x) \leq i(x)$
\I
For all $i$, $\cexpct{\pi'}{X'_i}  \leq 1 + \cexpct{\pi}{X_i}$
\EE
Together they give what we need:
\[
\cexpct{\pi'}{X'_{i'(x)}} \leq 
1 + \cexpct{\pi}{X_{i'(x)}} \leq 1 + \cexpct{\pi}{X_{i(x)}}
\]
The first is actually true trivially for all sequences, as the sooner the first door opens, the better the expected time to finish. For the second,
since for all $i$, $\pi'_i \geq \pi_i$, then $x \geq \pi'_i$ implies that $x \geq \pi_i$, and so:
\[
i'(x) = \max_i\set{ x \geq \pi'_{i-1}} \leq \max_i\set{ x \geq \pi_{i-1}} = i(x)
\]
For the third, 
denote by $Y_j$ the event that door 2 opens at the $j$'th 2-knock. Then:
\[
\cexpct{\pi}{X_i} 
= \sum_{j=i}^\infty (\pi_j + cj) \cprob{Y_j}{X_i}
\]
Let us consider this same expression as it occurs in $\pi'$. First note that 
$\cprob{Y_j}{X_i} = \cprob{Y_j'}{X_i'}$, as all that matters for its evaluation is $j-i$. Therefore: 
\[\BA
\cexpct{\pi'}{X'_i} 
& = 
\sum_{j=i}^\infty (\pi'_j + cj) \cprob{Y'_j}{X'_i}
\leq
\sum_{j=i}^\infty (\pi_j + 1 + cj) \cprob{Y_j}{X_i}
\\ & =
\cexpct{\pi}{X_i} + \sum_{j=i}^\infty \cprob{Y_j}{X_i}
\leq 
\cexpct{\pi}{X_i} + 1
\EA\]
\EPF

\subsection{The Optimal Semi-Fractional Sequence}

A big advantage of the semi-fractional model is that we can find an optimal sequence
for it. For that we need  some preparation:
\begin{definition}
For a semi-fractional sequence $\pi$, and some $0 \leq x \leq 1$,  denote by $\sexpct{x}{\pi}$ the expected running time of $\pi$ when started with door 1 being closed with probability $x$. In this notation, $\expct{\pi} = \sexpct{1}{\pi}$.
\end{definition}

\begin{lemma} \label{lm:x}
Let $y = x/(q_2 + p_2x)$. Then:
\[
\quad
\sexpct{x}{1^t \cdot \pi} = t + \sexpct{q_1^tx}{\pi} 
\quad \quad \quad \quad \quad
\sexpct{x}{2 \cdot \pi} = c + \F{x}{y}\sexpct{y}{\pi}  
\]
\end{lemma}
\BPF
The first equation is clear, since starting with door 1 being closed with probability $x$, and then knocking on it for $t$ rounds, the probability that this door is closed is~$q_1^t x$.

As for the second equation, if door 1 is closed with probability $x$, then knocking on door~2, we have a probability of $p_2(1-x)$ of terminating, and so the probability we did not finish~is:
\[
1 - p_2(1-x) = 1 - p_2 + p_2x = q_2 + p_2x = \F{x}{y}
\]
It remains to show that conditioning on the fact that we indeed continue, the probability that door 1 is closed is $y$.
It is the following expression, evaluated after a 2-knock:
\[
\F{\prob{\text{door 1 is closed}}}
{\prob{\text{door 1 is closed}} + \prob{\text{door 1 is open but not door 2}}}
=
\F{x}{x + (1-x)q_2} 
=
y 
\]
\EPF
Applying Lemma \ref{lm:x} iteratively on a finite sequence $w$, we get:
\[[eq:state]
\sexpct{x}{w \pi} = a(x,w) + b(x,w) \sexpct{\state(x,w)}{\pi}
\]
Of specific interest is $\state(x,w)$. It can be thought of as the {\em state}\footnote{There is an intuitive meaning behind this. Going through Lemma \ref{lm:x}, we can see that $\state(1,w)$ is actually the probability that after running $w$, door 1 is closed conditioned on door 2 being closed. Indeed, After running some finite sequence, the only feedback we have is that the algorithm did not finish yet. We can therefore calculate from our previous moves what is the probability that door 1 is closed, and that is the only information we need for our next steps.}
of our algorithm after
running the sequence $w$, when we started at state $x$. 
Lemma \ref{lm:x} and Equation \eqref{eq:state} give us the behaviour of~$\state(x,w)$:
\begin{flalign*}
\quad
\state(x,1^t) &= q_1^t x 
& 
\state(x,2) & = \F{x}{q_2 + p_2 x} 
& 
\state(x,a w) & = \state(\state(x,a), w)
\quad
\end{flalign*}
We start with the state being $1$, since we want to calculate $\sexpct{1}{\pi}$.
Except for this first moment, as we can safely assume any reasonable algorithm will start with a 1-knock,
the state will always be in the interval $(0,1)$. A 1-knock will always decrease the state and a 2-knock will increase it.

Our point in all this, is that we wish to exploit the fact that our doors are memoryless, and if we encounter a state we've already been at during the running of the sequence, then we should probably make the same choice now as we did then. The following definition and lemma capture this point.
\begin{definition}
We say a non-empty finite sequence $w$ is $x$-invariant, if $\state(x,w) = x$.
\end{definition}
The following Lemma is proved in Appendix \ref{app:improve}, and formalizes our intuition about how an optimal algorithm should behave.
\rLemma{lmImprove}{
If $w$ is $x$-invariant, and $\sexpct{x}{w\pi} \leq \sexpct{x}{\pi}$ then 
$\sexpct{x}{w^\infty} \leq \sexpct{x}{w\pi}$.
}

\subsubsection{The Actual Semi-Fractional Sequence}

\begin{theorem} \label{thm:optimal2}
There is an optimal semi-fractional sequence $\pi^\star$ of the form $1^s (2 1^t)^\infty$, for some positive real values $s$ and $t$, and its running time is:
\[
\expct{\pi^\star}  = \min_{z \in [0,1]}\B{ \log_{q_1}(1 - z) + \F{c + (1-p_2z)\log_{q_1}(1 - p_2z)}{p_2z} }
\]
\end{theorem}
\BPF
Claim \ref{clm:optimal2fractional} of Appendix \ref{sec:optimal-fraction} says that there is an optimal semi-fractional sequence $\pi$.
It clearly starts with a non-zero 1-knock, and so we can write $\pi = 1^s 2 \pi'$.
Intuitively, in terms of its state, this sequence starts at $1$, goes down for some time with a 1-knock, and then jumps back up with a 2-knock. The state it reaches now was already passed through on the first 1-knock, and so as this is an optimal sequence we can assume it will choose the same as it did before, and keep zig-zaging up and down. 

\begin{figure}[ht]
\includegraphics[scale=0.6]{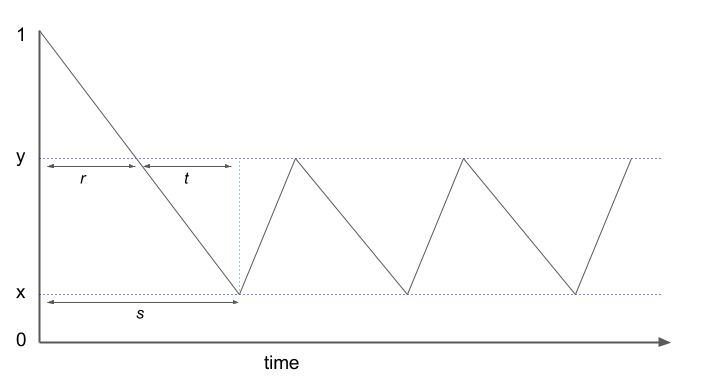}
\centering
\caption{\small How the state evolves as a function of time. 1-knocks decrease the state, and 2-knocks increase it. Note that $r = \log_{q_1}(y)$ and $s = \log_{q_1}(x)$.}
\label{fig:forProof}
\end{figure}

We next prove that indeed there is an optimal sequence following the zig-zaging form above.
Again, take some optimal $\pi$, and write $\pi = 1^s 2 \pi'$.
Denote $x = \state(1, 1^s)$ and $y = \state(1, 1^s 2) = \state(x, 2) > x$ (see Figure \ref{fig:forProof}).
Taking $r = \log_{q_1}(y) < s$, we get $\state(1, 1^r) = y$. Denoting $t=s-r$, this means that $1^t 2$ is $y$-invariant.
Since $\pi$ is optimal, then:
\[
\expct{\pi} = \expct{1^r (1^t 2) \pi'} \leq \expct{1^r \pi'}
\quad \text{ which implies: } \quad
\sexpct{y}{1^t 2 \pi'} \leq \sexpct{y}{\pi'}
\]
So by Lemma \ref{lmImprove}:
\[
\sexpct{y}{(1^t  2)^\infty} \leq \sexpct{y}{1^t 2 \pi'}
\quad \text{ which implies: } \quad
\expct{1^r(1^t 2)^\infty} \leq \expct{1^r 1^t 2 \pi'} = \expct{\pi}
\]
Therefore, $1^r (1^t 2)^\infty = 1^s(2 1^t)^\infty$ is optimal. We denote this sequence $\pi^\star$.

Now for the analysis of the running time of this optimal sequence. We will use Lemma \ref{lm:x} many times in what follows.
\[
\sexpct{1}{1^s (2 1^t)^\infty} = s + \sexpct{x}{(2 1^t)^\infty}
\]
Denote $\alpha = (2 1^t)^\infty$.
\[
\sexpct{x}{\alpha} = \sexpct{x}{2 1^t \alpha} 
= c + \F{x}{y}\sexpct{y}{1^t\alpha}
= c + \F{x}{y}(t + \sexpct{x}{\alpha})
\]
Since $t = s- r = \log_{q_1}(x/y)$:
\[
\sexpct{x}{\alpha} 
= \F{c}{1 - \F{x}{y}} + \F{\F{x}{y}}{1 - \F{x}{y}}\log_{q_1}(x/y)
\]
By Lemma \ref{lm:x}, as our $y$ is the state resulting from a 2-knock starting at state $x$, it follows that $y = x/(q_2 + p_2x)$.
Since $x/y = q_2 + p_2x$, then $1 - x/y = p_2(1-x)$ and then we get:
\[
\F{c}{p_2(1-x)} + \F{q_2 + p_2x}{p_2(1 - x)}\log_{q_1}(q_2 + p_2x)
\]
And in total:
\[
\sexpct{1}{1^s (2 1^t)^\infty} = 
\log_{q_1}(x) + \F{c + (q_2 + p_2x)\log_{q_1}(q_2 + p_2x)}{p_2(1-x)}
\]
Changing variable to $z=1-x$, results in $q_2 + p_2x = 1 - p_2z$, and we get the expression in the statement of the theorem.
\EPF

\subsection{Actual Numbers}\label{sec:approximating}

Theorem \ref{thm:optimal2} gives the optimal semi-fractional sequence and a formula to calculate its expected running time.
This formula can be approximated as accurately as we wish for any specific values of $p_1, p_2$ and $c$, but it is difficult to obtain a closed form formula from it.
Lemma \ref{lm:approx} in Appendix \ref{sec:lm:approx} gives us a pretty good result when $p_1 \approx p_2$, especially when they are small, as by Observation \ref{obsv:logq}, we get $\log(1/(1-p_1)) \approx p_1$, and so the additive mistake in the formula is something like $1$.

In general, when $p_1$ is small, then $\theta$ (see Lemma \ref{lm:approx})  is approximately $cp_1/p_2$, which is the expected time to open door 2 on its own, divided by the time to open door 1 on its own - a natural measure of the system. 
Then, ignoring the additive mistake, we get that the lower bound is approximately ${\cal F}(\theta)/p_1$, where $\cal F$ is some function not depending on the parameters of the system. For example ${\cal F}(1) = 3.58$. So opening two similar doors without feedback when $p$ is small takes about $3.58$ times more time than opening one door as opposed to the case with feedback, where the factor is only~$2$. 

We also note, that when the two doors are independent and similar, it is quite easy to see that the optimal expected running time is at most $3/p$ (see Claim \ref{claim:IID} in Appendix \ref{app:IID}). 
As a last interesting point, in Appendix \ref{apx:golden} we show that 
if $c=1$ and $p = p_1 = p_2$ approaches zero, then the ratio
between the number of 2-knocks and the number of 1-knocks approaches $\R{2}(1 + \sqrt{5})$, which is the golden ratio.

\subsection{Examples}

For $p_1=p_2=1/2$ and $c=1$, the lower bound is $5.747$. Simulations show that the best algorithm for this case is slightly more than $5.8$, so the lower bound is quite tight, but our upper bound is $6.747$ which is pretty far. 
However, the sequence we get from the upper bound proof starts with:
\[
1, 1, 2, 1, 2, 2, 1, 2, 1, 2, 2, 1, 2, 2, 1, 2, 1, 2, 2, 1, 2, 1, 2, 2, 1, 2, 2, 1, 2, 1, 2, \ldots
\]
The value it gives is about $5.832$, which is very close to optimal.

For $p_1=p_2=1/100$ and $c=1$, the sequence we get is:
\[
1^{97}, 2, 2, 1, 2, 2, 1, 2, 1, 2, 2, 1, 2, 1, 2, 2, 1, 2, 2, 1, 2, 1, 2, 2, 1, 2, 2, 1, 2, 1, 2,
\ldots
\]
And the value it gives is about $356.756$, while the lower bound can be calculated to be approximately $356.754$. As we see this is much tighter than the $+1$ that our upper bound promises.


\clearpage

\appendix

\section{``Real Life'' Example} \label{app:example}

The following scenario, much simplified, is the focus of \cite{FOCS}. A {\em treasure} $\tau$ is placed at a leaf of a $\Delta$-regular rooted tree of depth $d$. A mobile agent starting at the root wishes to find it as fast as possible and is allowed to move along edges. At each node there is an {\em advice} pointer to one of its neighbours. With probability $p$ the advice is {\em correct}, {\em i.e.,} directed towards the treasure, and with probability $q=1-p$ it is {\em incorrect}, {\em i.e.,}  points towards the one of the neighbours uniformly at random. The agent can move between two neighbouring nodes in one unit of time and look at the advice at the current node hosting it.
Minimizing the expected running time  until finding the treasure turns out to be not trivial, and the crux of the problem is that the advice is permanent, and so cannot be amplified by rechecking it. It is shown in \cite{FOCS} that if $q \gtrsim 1/\sqrt\Delta$, then no algorithm has  running time which is polynomial in $d$. Modeling this problem as cascading doors gives a non-trivial solution for the cases where $q$ is smaller. 

Consider door $i$ as open once the algorithm visits $\tau_i$, the ancestor  of the treasure that is at distance $i$ from the root. The purpose is then to open all doors. 
At any point in the algorithm, the \emph{candidates} at level $i$ are those unvisited vertices at that level, whose parent is already visited.
Also denote the score of a vertex as the number of advice pointers that point towards it in the advice seen so far by the agent. A knock on door $i$ consists of visiting the highest scoring candidate on level $i$, where symmetry is broken arbitrarily.

The difference in score between two candidates at the same level is affected by the advice on the path between them only, and as the algorithm moves on edges, all of the advice on this path is known.  
Consider a candidate at level $i$ that is at distance $l$ from $\tau_i$. 
It will have a score that is at least as high as the treasure if the number of advice pointers on the path connecting them that point towards it is greater than the number of those pointing towards $\tau_i$.
The probability that this happens can be viewed as the probability that a random walk of length $l$ sums up to at least $0$, where each step is $(-1,0,1)$ with respective probabilities $(p + q/\Delta, (1-2/\Delta)q, q/\Delta)$. Denote this probability by $\alpha(l)$. It is shown in \cite{FOCS} that $\alpha(l) \approx q^l$ for $q < 1/\sqrt{\Delta}$.

Denote by $C_i$ the number of such candidates at level $i$ that ``beat'' $\tau_i$. Even assuming all of the vertices at that level are now reachable:
\[
\expct{C_i} \leq \sum_{j=1}^i \Delta^j \alpha(2j-1) \approx \R{q} \sum_{j=1}^i (\Delta q^2)^j = O(\sqrt\Delta)
\]
This is in fact an upper bound on the expected number of knocks until opening door $i$, assuming door $i-1$ is already open. By Example \ref{exMoment},  $\simple$ will need an expected $O(d^2 \sqrt\Delta)$ knocks to open all doors and find the treasure. 
As moving from one candidate to another takes $O(d)$ moves, the running time of this algorithm is at most $O(d^3 \sqrt\Delta)$. If we were able to prove that $\expct{C_i^2}$ is small, then by Example \ref{exMoment} we could have dropped $d$'s exponent to $2.5$, but it turns out that this second moment is actually exponential in $d$ and so this approach fails.

Of course, assuming there is no feedback at all in this situation is an over approximation, and while it gives a non-trivial result, using much more sophisticated arguments, it is shown in \cite{FOCS} that there is an $O(\sqrt{\Delta} d)$ algorithm, and that it is in fact optimal.

\section{General Dependencies} \label{apx:general}

In the main text of the paper we focus on two special cases, that of independent doors and that of cascading doors. In what follows we introduce the possibility of much more general dependencies, and show that in fact, the two cases above are the extreme ones and so proving their equivalence is enough to prove it for all cases. For that we need to revisit our basic definition of doors and knocks.

\subparagraph*{Acyclic dependencies.} We assume that the directed graph of dependencies between doors is {\em acyclic}. In such cases, the doors can be ordered in a topological order such that a door may depend only on lower index doors. In what follows, w.l.o.g., we shall always assume that doors are ordered in such an order.

\subparagraph*{Configuration.}
A configuration $\cC$ for $d$ doors indexed $1, \ldots, d$ describes the probabilities of each door opening as a result of knocks on it. It relates a door $i$ with the function:
\[
\phi^\cC_i: \stset{ \B{X_1, \ldots, X_n}}{ n \geq 1, \forall j. X_j \subseteq \Ri} \rightarrow [0,1]
\]
Which given, for a sequence of $n$ knocks on door $i$, the set of doors $X_j$ that are open at the time of each of those knocks, returns the probability that door $i$ was opened by one of these knocks.
We will omit the superscript $\cC$ when it is clear from context.

\subparagraph*{Monotonicity.}
The more we knock on a door the better our chances of opening it. More precisely, the {\em monotonicity property} requires that if $X = (X_1, \ldots, X_n)$, and $X'$ is a sub-sequence of $X$ (possibly a non-consecutive one),
then $\phi_i(X') \leq \phi_i(X)$.

\subparagraph*{Positive Correlation.} 
We focus on the case where the doors are {\em positively correlated}, namely, a door being open can never decrease the chances of other doors to open. Formally this means that for every $i$, 
if for all $j$, $X_j' \subseteq X_j$, then $\phi_i(X_1',\ldots,X_n') \leq \phi_i(X_1,\ldots,X_n)$.

\subparagraph*{Fundamental Distribution.}
The {\em fundamental distribution}\footnote{This is actually not a distribution function, but rather the complement of an accumulative distribution function.} of door $i$ in configuration $\cC$ is the function $p^\cC_i$ (again, we will omit the superscript) where
$p_i(n)$ denotes the probability, in the best of conditions, {\em i.e.}, when all doors it depends on are open, that door $i$ remains closed after being knocked on $n$ times. Formally,  
$p_i(n) = 1 - \phi_i(\Ri^n)$. 
So $p_i(0) = 1$ for every door $i$, and by the monotonicity property $p_i$ is non-increasing. We also denote by $E_i = \sum_{n=0}^\infty p_i(n)$ the expected 
time to open door $i$ assuming all the doors it depends on are already open. We will always assume that for all $i$, $E_i < \infty$.

\subparagraph*{Similarity.}
Two doors are {\em similar} if they have the same fundamental distribution.
Two configurations are {\em similar} if for every $i$, door $i$ of the first configuration is similar to door $i$ of the second.

\subsection{The Cascading and Independent Configurations} \label{sec:general}

In light of the definitions above we define the two main configurations:
\BE
\I {\em Independent doors.} The distribution associated with a door is independent of whether or not other doors are open. 
Formally, $\phi_i(X_1, \ldots, X_k) = \phi_i(\Ri^k)$. 
\I  {\em Cascading doors.} Door $i>1$ cannot open unless door $i-1$ is already open. Only after door $i-1$ opens we start counting knocks on door $i$. 
Formally, $\phi_i(X_1, \ldots, X_k) = \phi_i(\Ri^t)$ where $t$ is the number of $X_j$'s that are  equal to $\Ri$. 
\EE

\begin{definition} \label{def:dominates}
For configurations $\cA$ and $\cB$,
we say that $\cA$ dominates $\cB$, if for every $i$ and every $X = (X_1, \ldots, X_n)$, we have: 
$\phi^{\cA}_i(X) \geq \phi^{\cB}_i(X)$.
\end{definition}
First:
\begin{claim}
For configuration $\cC$, similar independent configuration $\cI$, and similar 
cascading configuration $\cX$, 
$\cI$ dominates $\cC$ and $\cC$ dominates $\cX$.
\end{claim}
\BPF
Denote $n = |X|$, and denote by $k$ the number of elements of $X$ that are equal to $\Ri$.
We get the following series of inequalities:
\[\BA
\phi_i^\cX(X) 
& = 
1 - p_i(k) 
=  
\phi_i^\cC(\Ri^k) 
\\ & \leq 
\phi_i^\cC(X)
\leq 
\phi_i^\cC(\Ri^n) 
= 
1 - p_i(n)
= 
\phi_i^\cI(X)
\EA\]
Where we used, in order: the definition of cascading configuration, the fact that $p_i$ is the fundamental distribution of door $i$, monotonicity, positive correlation, the fact that $p$ is the fundamental distribution of door $i$, and the definition of independent configuration.
\EPF
An important property of dominance is:
\begin{claim} \label{clm:dominance}
For any sequence $\pi$,
if $\cA$ dominates $\cB$ then $\Time_{\cA}(\pi) \leq \Time_{\cB}(\pi)$.
\end{claim}
\BPF 
A possible way to describe the random process governing the running of an algorithm in a particular configuration, is as follows:
\BE
\I
For each door $i$, choose uniformly at random a real number $a_i \in [0,1]$. Fix this number for the rest of the run.
\I
Denote by $X$ the history of open doors as usual. Start it as the empty sequence.
\I
Go over the knocks in the sequence in order, and when the knock is on door $i$, check if 
$\phi_i(X') > a_i$, where $X'$ is part of $X$ that is relevant to calculate $\phi_i$ (only the indices where there is a knock of door $i$, and only the information about the doors of $\Ri$). If it is then consider door $i$ as open from this point on, and start marking it as such in $X$.
\EE 
This way of describing the run is a little bizarre, but is in fact very natural, as our doors are described by an accumulative distribution function. 

For two histories $X=(X_1, \ldots X_n)$ and $X'=(X'_1, \ldots X'_n)$,
we write $X' \preceq X$ if for all $j$, $X'_j \subseteq X_j$. 
We note that Definition \ref{def:dominates} combined with positive correlation, 
gives us that if $X' \preceq X$ then $\phi^{\cA}(X) \leq \phi^{\cB}(X')$.

This fact together with a simple  argument finishes the proof:
Use the same random coins to run the sequence $\pi$ in both $\cA$ and $\cB$.
By induction and the fact above, the histories at any point in time satisfy $X_\cB \preceq X_\cA$, and so the run on $\cA$ will always be at least as fast as the run on $\cB$. Since this is true no matter what $a_i$'s we got, it is true in expectation.
\EPF
Together these two claims prove the lemma we need for the paper:
\extreme

\subsection{A Simple Upper Bound on the Price of Lacking Feedback}\label{apx:upperPrice}

\begin{claim} \label{clm:upperPrice}
For every configuration $\cC$, $\Price(\cC) \leq d$.
\end{claim}
\BPF
Denote by $\cX$ the cascading configuration that is similar to $\cC$. Denote
$\pi = (1,2,\ldots,d)^\infty$.
Using Lemma \ref{extreme}, 
\[
\Time_\cC \leq \Time_\cC(\pi) \leq \Time_\cX(\pi)
\]
The behaviour of door $i$ in the cascading case can be described in a simple manner: It doesn't open until all lower index doors are open, and from that time is behaves according to $p_i$.  Hence, the expected number of knocks on door $i$ until it opens when starting the count after all doors $j<i$ are open, is precisely $E_i$.

In sequence $\pi$, it takes $d E_i$ to guarantee that door $i$ was knocked upon $E_i$ times.
Therefore, by linearity of expectation, it follows that the expected time until we open all doors in $\cX$ is at most $d\sum_{i=1}^d E_i$.
Dividing by $\sum_{i=1}^d E_i$, we get the result.
\EPF

\subsection{Existence of an Optimal Sequence}\label{apx:existence}

\begin{claim}\label{clm:optimal}
For any configuration $\cal C$
there is some sequence $\pi$ such that for every $\pi'$, 
$\Time_\cC(\pi) \leq \Time_\cC(\pi')$.
\end{claim}
\BPF
Assume there is no optimal sequence. 
Recall we assume that the fundamental distribution of each door allows it to be opened in  finite expected time. It is then easy to see that the sequence $(1,2,\ldots d)^\infty$ will open all doors in finite time no matter what the configuration is as long as it is acyclic. Therefore, $I = \inf_\pi \Time_\cC(\pi)$ exists.

Take a sequence of sequences $\pi^{(1)}, \pi^{(2)}, \ldots$ where
$\lim_{n\rightarrow \infty} \Time_\cC(\pi^{(n)}) = I$. W.l.o.g., we can assume that 
$\pi^{(n+1)}$  agrees with $\pi^{(n)}$ on all the first $n$ places. How so? there is at least one door number that appears as the first knock in infinitely many of the sequences. Take one such number, and erase all sequences that don't have it as a first knock. 
Of the remaining sequences, take the first one, fix it as $\alpha^{(1)}$, and erase it.
Starting with the sequence of sequences that remains, find a number that appears infinitely often in the second place. Erase all sequences not having it as the second knock, and then fix $\alpha^{(2)}$ as the first of the remaining sequences, erase it and continue thus. We get that the $\alpha^{(i)}$'s are a sub-sequence of the $\pi^{(i)}$'s, and satisfy the assumption.

Define $\pi_i = \lim_{n\rightarrow \infty} \pi^{(n)}_i$. It is clearly defined for such a sequence of sequences. This is our $\pi$.
Now:
\[\BA
\Time_\cC(\pi) 
& = 
\lim_{n\rightarrow \infty} \sum_{i=0}^n \prob{\pi \text{ not finished by time $i$}}
\\ & = 
\lim_{n\rightarrow \infty} \sum_{i=0}^n \prob{\pi^{(n)}\text{ not finished by time $i$}}
\\ & \leq
\lim_{n\rightarrow \infty} \Time_\cC(\pi^{(n)}) = I
\EA\]
Where the second equality is because $\pi = \pi^{(n)}$ in the first $n$ places.
\EPF

\section{Proofs Related to Section \ref{sec:multiple}} \label{apx:price}

\subsection{$\simple$ is Optimal up to a Constant Factor for Identical Independent Doors}\label{apx:clm:A1}
\clmAone
\BPF
By Claim \ref{clm:optimal}, there is some fixed sequence $\pi$ such that $\Time_\cI(\pi) = \Time_\cI$. Denote by $\pi_i(t)$ the number of times door $i$ has been knocked on by time $t$ in $\pi$. Clearly $\sum_i \pi_i(t) = t$.
\[\BA
\Time_\cI = \Time_\cI(\pi) 
& = 
\sum_{t=0}^\infty \prob{ \text{some door is closed at time $t$} }
\\ & =
\sum_{t=0}^\infty 1 - \prob{\text{all doors are open at time $t$} }
= 
\sum_{t=0}^\infty 1 - \left(\prod_{i=1}^d (1 - p(\pi_i(t)))\right)
\EA\]
By time $t$, the number of doors that have been tried more than $2t/d$ is less than $d/2$. So at least half the doors have been tried at most $t' = \Floor{2t/d}$ times. Therefore, each such door $i$ satisfies $p(\pi_i(t)) \geq p(t')$. We then have:
$\prod_{i=1}^d (1-p(\pi_i(t))) \leq \B{1-p(t')}^\F{d}{2}$. 
So $\Time_\cI$ is at least:
\[
\sum_{t=0}^\infty 1 - \B{1 - p\B{\Floor{{2t}/{d}}}}^\F{d}{2}
\]
In general, for any $x \leq 1$, as $t$ traverses all integers from $0$ to infinity, 
$\Floor{tx}$ takes every natural value at least $\Floor{1/x}$ times. In our case we get:
\[[eq:timeIbound]
\Time_\cI \geq \Floor{\F{d}{2}} \cdot \sum_{t=0}^\infty 1 - (1-p(t))^\F{d}{2}
\]
We now turn to analyse the expected running time of $\simple$. 
By Claim \ref{clm:simpleMax},  
$\Time_\cI(\simple) = O(d \cdot \expct{\max\B{X_1, \ldots, X_d}})$, where $X_i$ is the number of knocks on door $i$ until it opens.  
Now:
\[
\expct{\max\B{X_1, \ldots, X_d}} 
=
\sum_{t=0}^\infty 1 - \prob{X_i \leq t}^d 
= 
\sum_{t=0}^\infty 1 - (1-p(t))^d
\]
Denote $x(t) = (1-p(t))^{d/2}$, and then the sum $\sum_{t=0}^\infty (1 - (1-p(t))^d)$ becomes:
\[
\sum_{t=0}^\infty 1 - x(t)^2
=
\sum_{t=0}^\infty (1 - x(t))(1+x(t))
\leq
2\sum_{t=0}^\infty 1 - x(t) = 2\sum_{t=0}^\infty 1 - (1 - p(t))^{d/2}
\]
Applying this, and then using Equation \eqref{eq:timeIbound} we get:
\[
\Time_\cI(\simple) 
= 
O\B{d\sum_{t=0}^\infty 1 - (1 - p(t))^{d/2}}
= 
O(\Time_\cI) 
\]
\EPF

\subsection{A Simple Algorithm for General Configurations Where all Doors are Similar}\label{apx:simpleX}

\clmSimpleX
\BPF
Denote $\pi = \alpha_0 \cdot \alpha_1 \cdots$, and note that $|\alpha_n| = 2^n d$.
By Lemma \ref{extreme} we need only consider $\Time_\cX(\pi)$.
Taking $a_n = d+2d+\ldots+2^n d$, and using the right side of Observation \ref{obs:partial} (where indices are shifted to account for the fact that $a_0$ is the first element and not $a_1$):
\[\BA
\Time_\cX(\pi) 
& \leq
d + \sum_{n=0}^\infty 2^{n+1} d \cdot \nfin_\cX\B{\pi[d+2d+\ldots+2^n d)]}
\leq
d + 2\sum_{n=0}^\infty 2^n d \cdot \nfin_\cX(\alpha_n)
\\ & =
d + 4\sum_{n=0}^\infty 2^{n-1} d \cdot \nfin_\cI(\simple[2^n d])
\leq
d + 4\Time_\cI(\simple)
\EA\]
Where for the last step we used the left side of Observation \ref{obs:partial}, taking 
$a_n = 2^nd$.
Seeing as all doors start closed, $\Time_\cI(\simple) \geq d$, and we get that:
\[
\Time_\cC(\pi) = O(\Time_\cI(\simple)) = O(\Time_\cI) = O(\Time_\cC)
\]
Where for the last two steps, we used Claim \ref{clmAone}, and then Theorem \ref{extreme}.
\EPF

\subsection{Expected Maximum of iid Random Variables}\label{apx:IndTheta}

\lmMax
\BPF
Denote $p(n) = \prob{X_i > n}$.
The expectation we are interested in is:
\[[eq:twoTerms]
\sum_{t=0}^\infty 1 - \prob{X \leq t}^d
=
\sum_{t=0}^\infty 1 - (1 - p(t))^d
=
\sum_{t=0}^{\kappa-1} 1 - (1 - p(t))^d
+
\sum_{t=\kappa}^\infty 1 - (1 - p(t))^d
\] 
The first term is at least:
\[
\sum_{t=0}^{\kappa-1} 1 - \B{1 - \R{d}}^d \geq \kappa \B{1 - \R{e}}
\]
and at most $\kappa$, and so is $\Theta(\kappa)$.
For the second term, examine $(1 - a)^d$ when $a \leq 1/d$. We use $1+x\leq e^x$ and Observation \ref{obs:technical} (see below):
\[
(1 - a)^d \leq e^{-ad} \leq 1 - \R{2}ad
\]  
Hence the second term of \eqref{eq:twoTerms} is $\Omega(d\sum_{t = \kappa}^\infty p(t))$.
On the other hand, by the same observation:
\[
(1-a)^d \geq e^{-2ad} \geq 1 - 2ad
\]
And so the second term of \eqref{eq:twoTerms} is $O(d \sum_{t=\kappa}^\infty p(t))$.
\EPF

\begin{observation} \label{obs:technical}
Every $0\leq x \leq 1$ satisfies $e^{-x} \leq 1 - \R{2}x$.
\end{observation}
\BPF
Define:
\[ 
f(x) = 1 - \R{2}{x} - e^{-x}
\]
Whenever $f$ is positive the required inequality is satisfied.
We note that $f(0) = 0$ and $f(1) = 1 - \R{2} - \R{e} > 0$.
Now,
\[
f'(x) = -\R{2} + e^{-x} 
\]
It is positive for $x < \ln(2) < 1$, zero at $\ln(2)$, and negative for larger values. So $f$ starts as $0$ at $0$, climbs up to reach its maximum at $\ln(2)$ and then decreases. Since $f(1) > 0$, it must be the case that for all $0 \leq x \leq 1$, $f$ is positive, which proves the lemma.
\EPF

\subsection{Proofs for the Examples of Subsection \ref{sub:price}} \label{apx:examples}

\exGeometric
\BPF
In this case, $\kappa = \lceil\log_{1/q} (d)\rceil$, and $\expct{X_i} = 1/(1-q)$, so by \eqref{eq:priceByMax}:
\[
\Price = \Theta \B{ (1-q)\Ceil{\log_{1/{q}}(d)} + d(1-q)\sum_{i=\kappa}^\infty q^i}
\]
The second term inside the brackets is equal to $d q^\kappa \leq 1$. The first term 
is at least:
\[
\F{1-q}{\ln(1/q)}\ln(d) \geq q \ln(d) \geq \R{2} \ln(d)
\]
Where we used Observation \ref{obsv:logq} below.
On the other hand, it is at most:
\[
(1-q)\Ceil{\F{\ln(d)}{1-q}} 
\leq 
(1-q)\B{1 + \F{\ln(d)}{1-q}}
\leq 1 + \ln(d) \leq 3\ln(d)
\]
Since $d \geq 2$ and so $2\ln(d) > 1$.
So we get the result.
\EPF

\begin{observation} \label{obsv:logq}
For $0<q<1$,
\[
1-q \leq \ln\BF1q \leq \F{1-q}{q}
\]
\end{observation}
\BPF
The following is true for all $x > -1$:
\[
\F{x}{1+x} \leq \ln(1+x) \leq x
\]
So for $0<x<1$:
\[
\F{-x}{1-x} \leq \ln(1-x) \leq -x
\]
Which is:
\[
x \leq \ln\BF{1}{1-x} \leq \F{x}{1-x}
\]
Taking $x=1-q$ we get the first result.
\EPF

\exPolynomial
\BPF
In this case $\kappa = \Ceil{(dc)^{1/a}}$.
In this proof we have many approximations (such as dropping the rounding above), and they all go into the constants. 

The expected time to open just one door is (we assume $c^\R{a}$ is an integer, again this will only cost a constant factor):
\[
\sum_{n=0}^\infty p(n) 
=
c^\R{a} + \sum_{n=c^{1/a}}^\infty \F{c}{n^a}
\approx 
c^\R{a} + c\int_{c^{1/a}}^\infty \R{x^a} \diff x
\approx 
c^\R{a} + \F{c}{(a-1)c^{1-\R{a}}}
=
c^\R{a} \B{1 + \R{a-1}}
\]
On the other hand, in the terminology of Equation \eqref{eq:priceByMax}:
\[
\expct{\max\set{X_1, \ldots, X_d}} = 
\Theta\B{
(dc)^\R{a} + d\sum_{i=\kappa}^\infty p(i)
}
\]
We approximate the sum in second term in the brackets by an integral:
\[
d\sum_{i=\kappa}^\infty \F{c}{i^a} \approx
d\int_{\kappa}^\infty \F{c}{x^a} \diff x =
\F{dc}{(a-1)\kappa^{a-1}} \approx
\F{dc}{(a-1)(dc)^{1 - \R{a}}}
=
\F{d^\R{a} c^\R{a}}{a-1}
\]
So the expectation of the maximum is:
\[
d^\R{a} c^\R{a} \B{1 + \R{a-1}}
\]
And we get the result.
\EPF

\exMoment
\BPF
\[
p(n) = \prob{X_1 > n} = 
\prob{X_1^a > n^a} < \F{\expct{X_1^a}}{n^a} \leq \F{M}{n^a}
\]
So the current configuration dominates the independent door configuration where each door has fundamental distribution $q(n) = M/n^a$, and so by Claim \ref{clm:dominance} has algorithms with running time at least as good. 
Following the proof of Example \ref{exPolynomial}, there is such an algorithm with running time 
$O\B{d^{1+\R{a}} M^\R{a} \B{1 + \R{a-1}}}$.
\EPF

\section{Proofs Related to Section \ref{sec:geometric}}

\subsection{The Existence of an Optimal Semi-Fractional Sequence} \label{sec:optimal-fraction}

\begin{claim} \label{clm:optimal2fractional}
There is an optimal semi-fractional sequence $\pi$. That is, for every semi-fractional sequence $\pi'$, 
$\expct{\pi} \leq \expct{\pi'}$
\end{claim}
\BPF
Assume there is not.
But clearly, $I = \inf_\pi(\expct{\pi})$ exists. 
Take a series $\pi^1, \pi^2, \ldots$ where
$\lim_{n\rightarrow \infty} \expct{\pi^n} = I$. 

We think of a sequence as its sequence of 1-knock lengths. That is, $\pi^n_i$ is the length of the $i$-th 1-knock in $\pi^n$. 
We first show that we can assume 
that for every $i$, the set $\stset{ \pi^n_i }{n \geq 1}$ is bounded.

For this purpose, we first note that if for some semi-fractional sequence $\alpha$, $\expct{\alpha} < M$, then
for every $i$, $\alpha_i < M q_2^{i-1}$. That is because with probability at least $q_2^{i-1}$ the algorithm will actually run the $i$-th 1-knock, and if it's longer than stated, then $\expct{\alpha} \geq M$, in contradiction.
Since we can assume that for all $n$, $\expct{\pi^n} < 2I$, then by the observation above, we get the boundedness property we were aiming for.

Now, we claim that we can assume that for every $i$, $\pi^n_i$ converges as $n$ goes to infinity.
For this, start by taking a sub-series of the $\pi^n$ where $\pi^n_1$ converges (it exists, because these values are bounded, as we said). Erase all other $\pi^n$. Take the first element of this series and put aside as the new first element. From the rest, take a sub-series where $\pi^n_2$ converges, and erase all others. Take the new first element, and put it aside as the new second element. Continuing this, we get an infinite series as required. 

Define $\pi_i = \lim_{n\rightarrow \infty} \pi^n_i$. We claim that $\pi$ is optimal.
\[
\expct{\pi}
= 
\sum_{i=1}^\infty \B{\sum_{j=1}^i (\pi_j + c)} \prob{\text{$\pi$ finishes at 2-knock $i$}}
\]
Denoting by $X_i$ the event that $\pi$ finishes at or after 2-knock $i$, this is equal to:
\[
\sum_{i=1}^\infty (\pi_i + c) \prob{X_i}
= \lim_{k \rightarrow \infty} \sum_{i=1}^k (\pi_i + c) \prob{X_i}
\]
Fix some $k$. And denote by $X^n_i$ the event that $\pi^n$ finishes at or after 2-knock $i$.
Since $\prob{X_i}$ is a continuous function of $\pi_1, \ldots, \pi_i$, we get:
\[
\sum_{i=1}^k (\pi_i + c) \prob{X_i}
=
\lim_{n \rightarrow \infty} \sum_{i=1}^k (\pi^n_i + c) \prob{X^n_i}
\leq
\lim_{n \rightarrow \infty} \Time_\cI(\pi^n) = I
\]
So $\expct{\pi} \leq I$ and we conclude.
\EPF

\subsection{Memoryless Doors Imply Memoryless Algorithms}\label{app:improve}

\lmImprove
\BPF
As in \eqref{eq:state}, for any sequence $\alpha$:
\[
\sexpct{x}{w \alpha} = a + b \sexpct{x}{\alpha}
\]
Where $a$ and $b$ are functions of $x$ and $w$.
Since $w$ is not empty, and as 1-knocks decrease the state 
and 2-knocks increase it, there must be at least one 2-knock in $w$, and thus $b < 1$.
So:
\[
\sexpct{x}{w \pi} = a + b \sexpct{x}{\pi}
\geq
a + b \sexpct{x}{w \pi}
\RIGHT
\sexpct{x}{w \pi} \geq \F{a}{1-b}
\]
On the other hand:
\[
\sexpct{x}{w^\infty} = \sexpct{x}{w w^\infty} = a + b \sexpct{x}{w^\infty}
\RIGHT
\sexpct{x}{w^\infty} = \F{a}{1-b}
\]
And we conclude.
\EPF

\subsection{Approximating the Optimal Semi-Fractional Running Time}\label{sec:lm:approx}

Theorem \ref{thm:optimal2} gives a way to calculate the expectation of the best semi-fractional sequence $\pi^\star$ for our configuration. Unfortunately,
we were not able to obtain a close formula for this value. The following lemma can be used to approximate it.  
\begin{lemma} \label{lm:approx}
Denoting $\theta = -c\log(q_1)/p_2$, and 
$\psi = \R{2}(\sqrt{\theta^2 + 4\theta} - \theta)$, we have:
\[
\expct{\pi^\star}
\in
\R{\log(1/q_1)}\B{\log\BF1{1 - \psi} + \F{\theta}{\psi} + 1} 
- \left[0, \F{p_2}{\log(1/q_1)}\right]
\]
\end{lemma}
\BPF
Recall the result of Theorem \ref{thm:optimal2}:
\[
\expct{\pi^\star}  = 
\min_{z \in [0,1]}\B{ 
\log_{q_1}(1 - z) + \F{c +(1-p_2z)\log_{q_1}(1 - p_2z)}{p_2z}
} 
\]
By the definition of $\theta$ in the statement of the lemma, and denoting:
\[
Y = -\F{(1-p_2z)\log(1 - p_2z)}{p_2z}
\]
We get:
\[\BA
\expct{\pi^\star}  
&=
\R{\log(q_1)}
\min_{z \in [0,1]}\B{ 
\log(1 - z) - \F{\theta}{z} - Y
}
\\ & =
\R{\log(1/q_1)}
\min_{z \in [0,1]}\B{ 
\log\BF1{1 - z} + \F{\theta}{z} + Y
} 
\EA\]
Next, since for $x>-1$, 
\[
\F{x}{1+x} \leq \log(1+x) \leq x
\]
Then for $0<x<1$:
\[
-x \leq \log(1-x) \leq -\F{x}{1-x}
\]
Multiplying by $-(1-x)/x$ (a positive number):
\[
1-x \leq - \F{(1-x)\log(1-x)}{x} \leq  1 
\]
Therefore, $Y \in [1-p_2z, 1] \subseteq [1-p_2, 1]$. It follows that:
\[[eq:beforeMin]
\expct{\pi^\star}  \in 
\R{\log(1/q_1)}\B{
\min_{z \in [0,1]}\B{ 
\log\BF1{1 - z} + \F{\theta}{z}
} + [1-p_2, 1]}
\]
For the minimization, we take the derivative and compare to $0$
\[\BA
&
\R{1 - z} - \F{\theta}{z^2} = 0
\RIGHT
\F{z^2}\theta + z - 1 = 0
\\ & \RIGHT
z = \F{\sqrt{1 + 4/\theta} - 1}{2/\theta} = \F{\sqrt{\theta^2 + 4\theta} - \theta}2 = \psi
\EA\]
Where we took the root that is in $[0,1]$. Assigning back in \eqref{eq:beforeMin},
\[
\expct{\pi^\star}  \in 
\R{\log(1/q_1)}\B{\log\BF1{1 - \psi} + \F{\theta}{\psi} + 1} - \left[0, \F{p_2}{\log(1/q_1)}\right]
\]
\EPF

\subsection{Similar Independent Memoryless Doors}\label{app:IID}
The following simple claim implies that the expected time to open two similar memoryless doors is at most 3 times the expected time to open one of them. A generalization to $d$ doors can easily be established based on the same idea.
\begin{claim}\label{claim:IID}
Consider the configuration $\cI$ of two similar doors that open on each knock independently with probability~$p$. Then $\Time_\cI(\simple)=\F{3}p-1$.
\end{claim}
\BPF
Until the first door opens (either door 1 or door 2), each knock has probability $p$ to open. Therefore, the first door opens in expected time $1/p$. From that time, every odd knock will be on the other door, and  will succeed with probability $p$. So the expected time to open the second door after the first one has opened is $2/p-1$, and altogether we have expected time $3/p-1$. 
\EPF

\subsection{The Golden Ratio} \label{apx:golden}

Returning to the case where $c=1$, and $p_1 = p_2$ are very small. As we said, $\theta$ of Lemma \ref{lm:approx} tends to $1$, and so $\psi$ there tends to $(\sqrt{5}-1)/2$. This $\psi$ is actually the value of $z$ that minimizes the expression of Theorem \ref{thm:optimal2}. Looking in the proof of the theorem, the length of 1-knocks (except the first), is 
\[
t=\log_{q}(x/y) = \log_q(q + px) = \log_q(1 - pz) = \F{\log(1-pz)}{\log(1-p)}
\]
For small $x$, $\log(1+x) \approx x$ and so, as $p$ goes to zero, the above ratio tends to $z$, and in our case to $\psi$. 
So the length of 1-knocks is $\psi$, and that of the 2-knocks is $1$. In the long run the length of the first 1-knock is insignificant, and the transformation of Theorem \ref{thm:equiv2} will make the ratio of between the number of 2-knocks and the number of 1-knocks approach $1/\psi$, which is the golden ratio.

\clearpage

\bibliography{bib}

\end{document}